%% file: main.tex
\documentclass[letterpaper]{article} %
\usepackage{main}
\usepackage{times}  %
\usepackage{helvet}  %
\usepackage{courier}  %
\usepackage[hyphens]{url}  %
\usepackage{graphicx} %
\usepackage{natbib}  %
\usepackage{caption} %
\usepackage{algorithm}
\usepackage{algorithmic}
\usepackage{amsmath}
\usepackage{amsfonts}
\usepackage{booktabs}
\usepackage{tabularx}
\usepackage{multirow}
\usepackage{xcolor}
\usepackage[labelfont={small,bf},textfont={small}]{caption}
\usepackage{newfloat}
\usepackage{listings}
\DeclareCaptionStyle{ruled}{labelfont=normalfont,labelsep=colon,strut=off} %
\floatstyle{ruled}
\newfloat{listing}{tb}{lst}{}
\floatname{listing}{Listing}
\usepackage{amssymb}%
\usepackage{pifont}%

\title{MMP: Towards Robust Multi-Modal Learning with Masked Modality Projection}
\author{
    Niki Nezakati\textsuperscript{\rm 1},
    Md Kaykobad Reza\textsuperscript{\rm 1}, %
    Ameya Patil\textsuperscript{\rm 2},
    Mashhour Solh\textsuperscript{\rm 2},
    M. Salman Asif\textsuperscript{\rm 1}
}
\affiliations{
    \textsuperscript{\rm 1}University of California Riverside ~~~ 
    \textsuperscript{\rm 2}Amazon\\
}

\usepackage{tikz}

\nocopyright

\begin{document}

\maketitle

\begin{abstract}
Multimodal learning seeks to combine data from multiple input sources to enhance the performance of different downstream tasks. In real-world scenarios, performance can degrade substantially if some input modalities are missing. Existing methods that can handle missing modalities involve custom training or adaptation steps for each input modality combination. These approaches are either tied to specific modalities or become computationally expensive as the number of input modalities increases. In this paper, we propose \textbf{M}asked \textbf{M}odality \textbf{P}rojection (MMP), a method designed to train a single model that is robust to any missing modality scenario. We achieve this by randomly masking a subset of modalities during training and learning to project available input modalities to estimate the tokens for the masked modalities. This approach enables the model to effectively learn to leverage the information from the available modalities to compensate for the missing ones, enhancing missing modality robustness. We conduct a series of experiments with various baseline models and datasets to assess the effectiveness of this strategy. Experiments demonstrate that our approach improves robustness to different missing modality scenarios, outperforming existing methods designed for missing modalities or specific modality combinations.
\end{abstract}

\section{Introduction}
Multimodal learning (MML) \cite{baltruvsaitis2018multimodal, xu2023multimodal} leverages information from multiple input sources to perform better on the underlying task. Incorporating knowledge from diverse input sources has proven to be very effective in enhancing model performance \cite{huang2021makes, lu2024theory}. Since these models are generally trained on all input modalities, they rely heavily on the presence of all modalities to perform optimally during test time. In the real-world scenarios, any subset of the modalities can be missing due to sensor malfunction, privacy concerns, or data acquisition constraints. Recent studies have shown that multimodal models show significant performance drop when a subset of input modalities is missing \cite{ma2022multimodal, lee2023map}. In this paper, we investigate the missing modality issue during test time and show that a single model trained in a robust manner can outperform existing baseline methods in different missing modality scenarios.

\begin{figure}[t]
    \centering
    \includegraphics[width=1\columnwidth]{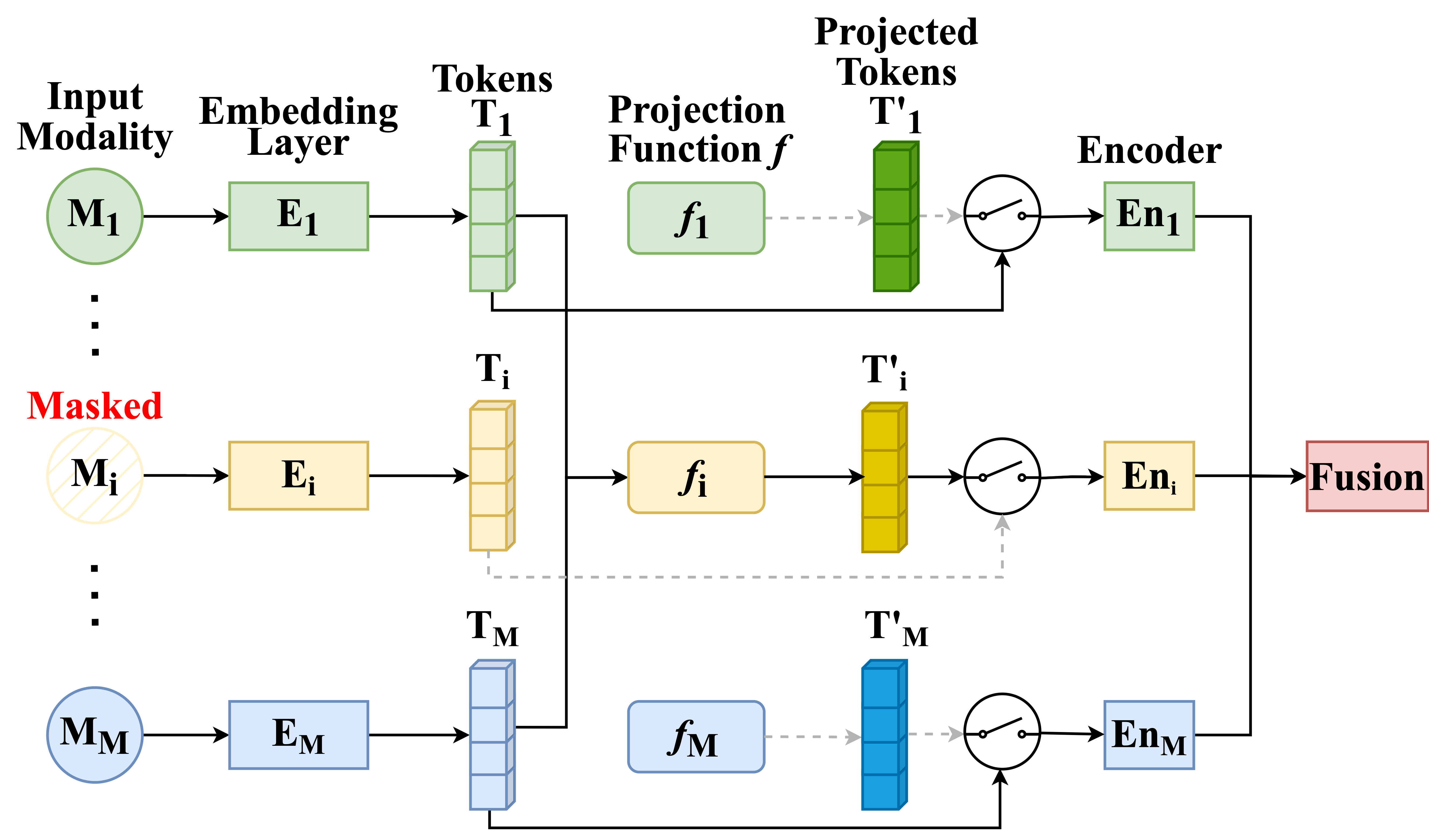} 
    \caption{Architecture of the proposed MMP approach for training a single multimodal model that is robust to missing modalities. Input modalities are passed through embedding layers, generating tokens. For a masked modality $i$, a projection function utilizes the tokens from the available modalities to generate projected tokens. These projected tokens are then passed to the masked modality branch.}
    \label{fig:overall-mmp-approach}
\end{figure}

A number of approaches has been proposed to enhance missing modality robustness for different multimodal tasks. Some of these approaches include utilizing a robust training approach \cite{neverova2015moddrop, hussen2020mdrop}, modality masking \cite{bachmann2022multimae, shin2023crm}, or knowledge distillation \cite{tarvainen2017meanteacher, maheshwari2023m3l, wu2024segment} during training. These approaches force the model to perform better with available modalities either by masking inputs or from the help of a teacher model. Prompting-based methods learn to compensate for the missing modalities with learnable prompts \cite{lee2023map, jang2024msp}; however, they need to learn one set of prompts for each missing modality scenarios, which scales poorly as the number of input modalities increase. Missing modality imputation is another approach where generative networks like generative adversarial network (GAN) \cite{yu2018_3Dcgan, sharma2019mri_gan} and variational autoencoder (VAE) \cite{dorent2019vae} are used to generate missing modalities from available modalities. Training such generative networks to impute missing modality adds extra overhead to the overall process.

In this paper, we propose \textbf{M}asked \textbf{M}odality \textbf{P}rojection (MMP), a method for training a single multimodal model that is robust to any missing modality scenario. Our approach leverages available modalities to compensate for the missing ones. In particular, we mask out a subset of modalities during training and utilize available modalities to predict the tokens for the masked ones. As illustrated in Figure~\ref{fig:overall-mmp-approach}, our approach consists of two main stages: (1) modality masking: where a subset of modalities is masked out during training; and (2) modality projection: where available modalities are used to predict the tokens of the masked modalities, referred to as projected tokens. Additionally, we use alignment loss to align the projected tokens and the actual tokens during training. During inference, the projected tokens are used to substitute for the missing modalities.

Our proposed MMP method can be integrated into any existing multimodal architecture. Moreover, we do not need to train or adapt the model for each missing modality scenario. Experimental results demonstrate that models trained with MMP show significant performance improvement in different missing modality scenarios. We conducted extensive experiments on three baseline models and five datasets, covering three tasks (Section \ref{section:experiments}). The results show that our method significantly improves performance when some modalities are entirely missing (compared to existing methods), while also maintaining competitive performance when all modalities are available. Performance of the MMP approach is also comparable or better than the models that are exclusively trained for each input modality combination.

Our main contributions can be summarized as follows.
\begin{itemize}
    \item Masked modality projection (MMP) is a novel approach to predict missing modality tokens from available modalities and enhance robustness to missing modalities.
    \item MMP provides significantly improved performance with missing modalities compared to models trained with all modalities. The performance is comparable or better than the networks  trained for specific modality combinations.
    \item MMP requires minimal change in the network, which makes it versatile and adaptable across various multimodal tasks, datasets, and models (as demonstrated by our experiments). 
\end{itemize}

\section{Related Work}
\label{section:related-work}
\textbf{Robust model design} is one approach to perform well on missing modality scenarios. \citet{wang2023shaspec} designed a model to learn modality specific and modality shared features and impute missing modality from available ones. On the other hand, \citet{shin2023crm} designed a robust framework based on modality masking and knowledge distillation for RGB-Thermal segmentation. \citet{wang2022multimodal_token_fusion} proposed a method to dynamically detect and replace uninformative tokens with informative tokens. \citet{Choi2019embracenet, fan2023spidermesh, lin2023vpfnet} designed a robust fusion strategies to enhance model robustness in different missing modality situations. For MRI
missing modality task, \citet{karimijafarbigloo2024mmcformer} proposed a  Transformer-based approach with adopted co-training strategy. However, these models and fusion strategies are generally designed for a specific tasks. So, it is non-trivial to generalize them for other multimodal tasks. 

\textbf{Robust training approach} can make models robust to missing modalities. Modality dropout augmentation has been applied by \citet{neverova2015moddrop} for multimodal gesture recognition and \citet{hussen2020mdrop} for generating 3D facial animation. Modality masking based approaches also gained popularity. \citet{shin2023crm} used complementary random masking and \citet{fan2023spidermesh} used partial modality masking for training robust models. \citet{bachmann2022multimae} utilized masked autoencoders, \citet{ma2022multimodal} utilized masked cross attention and \citet{hazarika2022analyzing} used modality perturbation to make the underlying models robust to missing and corrupted modalities. Though these approaches improve model robustness in missing modality scenarios, they can not compensate the performance drop completely. 

\textbf{Model adaptation} is another approach to make models robust to missing scenarios. \citet{lee2023map} trained one set of learnable prompts for each modality combination and used those learned prompts when modalities got missing during test time. A followup study by \citet{jang2024msp} showed that learning one set of prompts for each input modality is sufficient for comparable performance. \citet{reza2023robust} utilized parameter efficient adaptation to build a generic framework to make existing models robust. They validated the effectiveness of their framework on a number of multimodal tasks. The main disadvantage of these approaches is that they require one set of learnable parameters per modality combination.

\textbf{Generation and knowledge distillation} based approaches are also used to enhance model robustness. GAN based generative models were used by \citet{yu2018_3Dcgan}, \citet{sharma2019mri_gan}, and \citet{zhang2024unified}. \citet{dorent2019vae} used variational autoencoders to generate missing modality. \citet{qu2024lds2ae} introduced a local diffusion shared-specific autoencoder to the handle missing modality issue in image classification. Studies by \citet{woo2023actionmae} proposed ActionMAE to generate missing feature vectors for robust action recognition. Knowledge distillation also showed great promise in a number of tasks. \citet{shin2023crm} and \citet{maheshwari2023m3l} utilized knowledge distillation for multimodal segmentation tasks. \citet{wu2024segment} used a teacher-student distillation model to combine partially available visual information with auditory information. \citet{wei2023mmanet} presented a framework that uses a teacher network to transfer comprehensive multimodal information for improving multimodal learning with missing data. Apart from these approaches, policy learning \cite{ma2022multimodal}, Bayesian meta-learning \cite{ma2021smil} and weight-space ensembling \cite{wortsman2022wiseft} are also utilized for missing modality robustness. The main drawback of these approaches is that they need to train/utilize another model to generate missing modality or distill knowledge. 

In this paper, our goal is to train a single model that is robust to any missing modality scenario. Our method utilizes available input modalities to generate the tokens for the missing modalities. The model is trained end-to-end without the need for tuning or adapting it for any specific modality combination.

\section{Method}
In this section, we introduce Masked Modality Projection (MMP), a novel approach for training a single multimodal model that is robust to missing modalities. In this method, a subset of modalities is randomly masked out during each training iteration. To address the absence of these modalities, we introduce projection functions that learn to map tokens from the available modalities to the missing modality tokens, which we refer to as projected tokens. These projected tokens are aligned with the actual tokens using an alignment loss objective. Finally, the projected tokens are passed to the corresponding branch for the masked modality.

\subsection{Modality Masking}
Modality masking is a key component of our MMP approach. As discussed in Section \ref{section:related-work}, modality dropout augmentation, which randomly zeros out modalities, has shown robustness to missing modalities. 
We extend this idea by masking out all the tokens of a random subset of modalities during each training iteration. Instead of feeding zeros, we use available modalities to predict the tokens for the masked modalities. Specifically, at each iteration, a random subset of modalities are selected to be masked (i.e., they are not fed to the model). Our goal is to train the model to predict the tokens of the masked modalities using the available ones.

\begin{figure}[t]
    \centering
    \includegraphics[width=1\columnwidth]{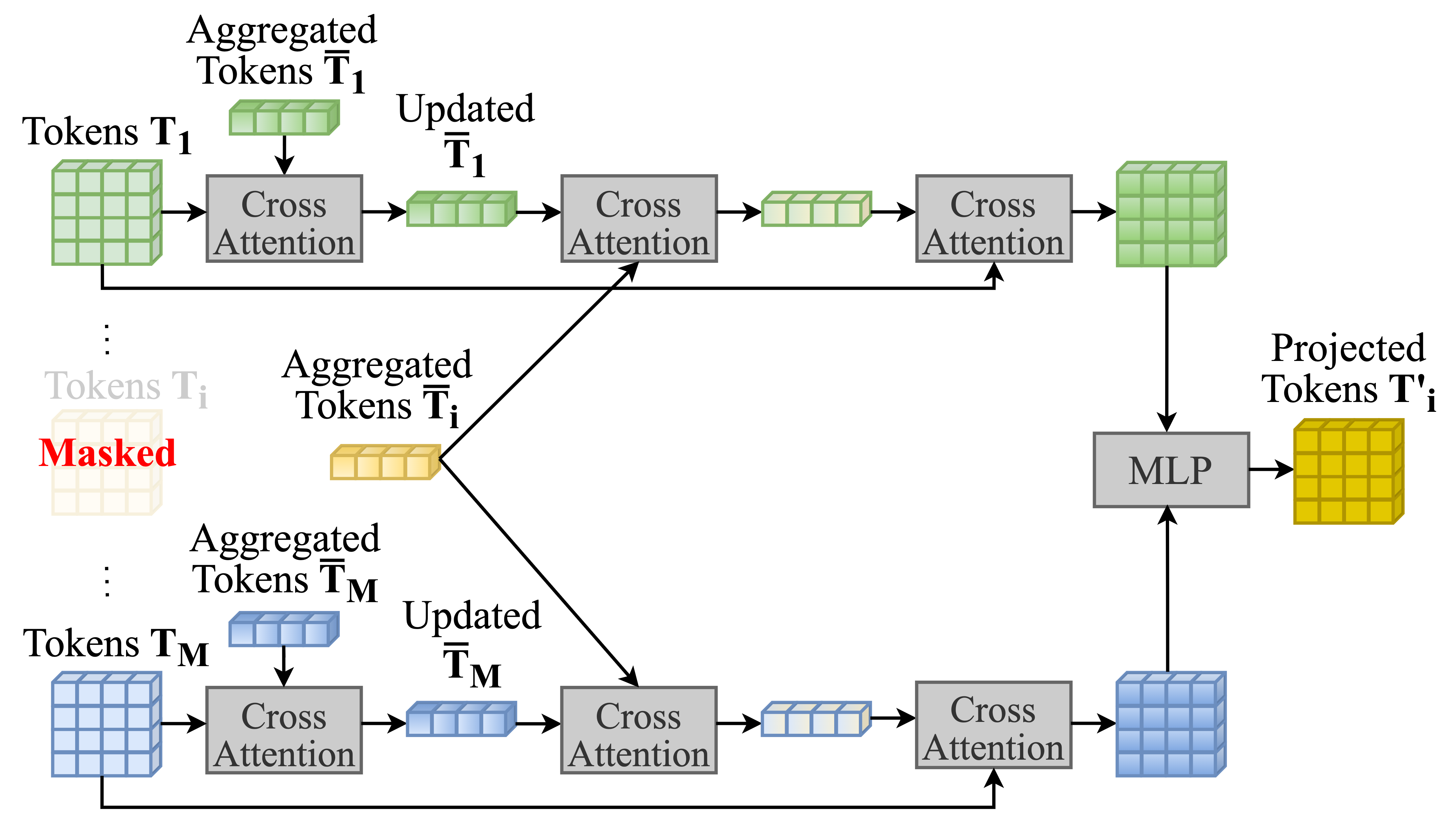} 
    \caption{Visualization of the modality projection approach. Available modality tokens are processed through cross-attention to update their aggregated tokens. These aggregated tokens are combined with those of the masked modality through another cross-attention step. The resulting cross-modal relationships are used to attend to the actual tokens of the available modalities. The final output tokens are passed through an MLP to generate the projected tokens of the masked modality.}
    \label{fig:projection-method}
\end{figure}

\subsection{Modality Projection}
We propose a modality projection approach for masked modality $i$, as illustrated in Figure \ref{fig:projection-method}. Suppose we have $M$ distinct modalities given as input. The embedding layers generate tokens from each input modality as 
\begin{equation} 
\mathbf{T}_i = \text{EmbeddingLayer}(\mathbf{I}_i),
\end{equation}
where $\mathbf{I}_i$ represents input modality $i\in\{1, 2, \ldots, M\}$ and $\mathbf{T}_i \in \mathbb{R}^{N \times d}$ denotes tokens for modality $i$, $N$ is the number of tokens, and $d$ is the embedding dimension. For simplicity, here we assume $N$ and $d$ to be the same across modalities. We will discuss how our method can be extended to handle cases with varying numbers of tokens and embedding dimensions among modalities in Section \ref{sec:method-dim-variability}.

We introduce the projection function $\textit{\textbf{f}}_i$ to predict the tokens of the masked modality $i$. This process begins by utilizing aggregated tokens. Inspired by \citet{mo2024unveiling}, aggregated tokens summarize the modality information into a compact representation; therefore, reducing the computational and storage complexity associated with numerous modality tokens. For each modality $i$, we use eight aggregated tokens $\mathbf{\overline{T}}_i$, initialized randomly as learnable parameters. When modality $j$ is available during an iteration, its aggregated tokens $\overline{\mathbf{T}}_j$ are updated by attending to the actual modality tokens $\mathbf{T}_j$ using multi-head cross-attention. This process is represented as
\begin{align}
\mathbf{\overline{T}}_j &= \text{CrossAttention}\left(\mathbf{\overline{T}}_j, \mathbf{T}_j \mid j \in \mathcal{A}\right), \\
& = \text{softmax}\left(\frac{\mathbf{\overline{T}}_j \mathbf{W}_\text{q} \mathbf{W}_\text{k}^\top\mathbf{T}_j^\top }{\sqrt{d}}\right) \mathbf{T}_j \mathbf{W}_\text{v}, \notag
\end{align}
where $\mathbf{T}_j \in \mathbb{R}^{N \times d}$ and $\mathbf{\overline{T}}_j \in \mathbb{R}^{8 \times d}$ denote tokens and aggregated tokens of modality $j$, respectively, and $\mathcal{A}$ is the set of available modalities. 

$\mathbf{W}_\text{q}$, $\mathbf{W}_\text{k}$, and $\mathbf{W}_\text{v}$ are learnable weight matrices for the query, key, and value projections, respectively. The aggregated tokens are only updated when a modality is available; if modality $i$ is masked at an iteration, its aggregated tokens $\mathbf{\overline{T}}_i$ are left unchanged. 

When dealing with missing modalities, for each modality $i$ that is missing (masked during training), cross-attention is performed between the aggregated tokens of the missing modality and the aggregated tokens of each available modality separately. This step captures the relationships between the missing modality and each available modality, allowing the model to approximate the missing modality information based on available data. Specifically
\begin{equation}
\mathbf{X}_{ij} = \text{CrossAttention}(\mathbf{\overline{T}}_i, \mathbf{\overline{T}}_j), 
\end{equation}
where $\mathbf{\overline{T}}_i, \mathbf{\overline{T}}_j$ are the aggregated tokens of missing modality $i$, available modality $j$, respectively. $\mathbf{X}_{ij}$ represents the attended tokens for available modality $j$ in relation to missing modality $i$.

We then utilize $\mathbf{X}_{ij}$ in cross-attention with the original tokens $\mathbf{T}_j$ of availble modality $j$ to ensure that the relational information is integrated with the specific features of each available modality. The refinement process is expressed as
\begin{equation}
\mathbf{T}_{\text{attended}_j} = \text{CrossAttention}(\mathbf{T}_j, \mathbf{X}_{ij}), 
\end{equation}
where $\mathbf{T}_{\text{attended}_j}$ denotes the refined attended tokens for available modality $j$. The refined attended tokens are then concatenated for each available modality $j$
\begin{equation}
    \mathbf{T}_{\text{available}} = \text{Concat}\left(\left\{\mathbf{T}_{\text{attended}_{j}} \mid j \in \mathcal{A}\right\}\right), 
\end{equation}
where $\mathcal{A}$ denotes the set of available modalities. The concatenated tokens \( \mathbf{T}_{\text{available}} \) are then fed into a multi-layer perceptron (MLP) to produce the projected tokens $\mathbf{T}'_i$ for the missing modality $i$ as 
\begin{equation}
\mathbf{T}'_i = \text{MLP}(\mathbf{T}_{\text{available}}). 
\end{equation}

The missing modality tokens are replaced with their corresponding projected tokens $\mathbf{T}'_i$ and passed to their respective branch in the network.

\subsection{Token and Dimension Variability}
\label{sec:method-dim-variability}
Our method can also be applied in cases where the number of tokens $N$ or the embedding dimension $d$ differs across modalities. To address differences in embedding dimensions, we apply a linear layer at the beginning of the projection process to map tokens from each modality to a common embedding dimension. For cases where the number of tokens varies, we incorporate a linear layer within the MLP module to align the token count of the projected tokens $\mathbf{T}'_i$ with that of the missing modality $i$.

\subsection{Alignment Loss Objective}
\label{sec:method-alignment}
To minimize the discrepancy between the projected and real tokens, we use an alignment loss objective to ensure that the projected tokens closely match their corresponding real tokens.  If $N_{\text{masked}}$ denotes the number of masked modalities at an iteration, the alignment loss is computed as
\begin{equation}
\mathcal{L}_{\text{alignment}} = \frac{1}{N_{\text{masked}}} \sum_{i \in \text{masked}} \mathcal{L}_{\text{alignment}_i}(\mathbf{T}'_i,\mathbf{T}_i),
\end{equation}
where $\mathcal{L}_{\text{alignment}_i}$ represents the Smooth L1 loss between the real tokens $\mathbf{T}_i$ and the projected tokens $\mathbf{T}'_i$. To incorporate the alignment loss into the overall optimization, we add $\mathcal{L}_{\text{alignment}}$ to the network's primary loss function as 
\begin{equation}
    \mathcal{L}_{\text{total}} = \mathcal{L}_{\text{task}} + \mathcal{L}_{\text{alignment}}
\end{equation}
where $\mathcal{L}_{\text{task}}$ represents the primary task-specific loss of the network.

\begin{table*}[t]
    \centering
    \begin{tabular}{@{}cccccc@{}}
    \toprule
    Dataset & Input Modalities & Missing Modalities & Pretrained & Modality Dropout & \textbf{MMP (ours)} \\ \midrule
    \multirow{4}{*}{MCubeS} & RGB-A-D-N   & -       & \textbf{51.54} & 48.56 & 48.95          \\
                            & A-D-N       & RGB     & 1.45           & 33.88 & \textbf{38.57} \\
                            & A-D         & RGB-N   & 0.93           & 33.15 & \textbf{37.74} \\
                            & A           & RGB-D-N & 1.13           & 26.3  & \textbf{31.31} \\ \midrule
    \multirow{3}{*}{NYUDv2} & RGB-Depth   & -       & \textbf{56.30} & 51.12 & 53.81          \\
                            & RGB         & Depth   & 51.05          & 48.80 & \textbf{52.04} \\
                            & Depth       & RGB     & 6.01           & 29.79 & \textbf{41.08} \\ \midrule
    \multirow{3}{*}{FMB}    & RGB-Thermal & -       & \textbf{62.68} & 54.11 & 60.03          \\
                            & RGB         & Thermal & 22.2           & 48.32 & \textbf{55.83} \\
                            & Thermal     & RGB     & 23.35          & 39.66 & \textbf{51.73} \\ \bottomrule
    \end{tabular}   
    \caption{Performance comparison (mIoU) of the pretrained model, modality dropout training, and MMP. Input and missing modalities columns indicate available and missing modalities during inference. A, D and N denote angle of linear polarization, degree of linear polarization, and near-infrared, respectively.}
    \label{tab:comparison-mcubes}
\end{table*}

\begin{table}[t]
    \centering
    \setlength{\tabcolsep}{3.5pt}
    \begin{tabular}{@{}llcccc@{}}
        \toprule
        \multicolumn{1}{c}{Methods} & \multicolumn{1}{c}{Backbone} & RGB & Depth & Avg. \\ \midrule
        AsymFusion \shortcite{wang2020learning_asym_fusion}  & ResNet-101 & 46.50          & 34.30          & 40.40          \\
        CEN \shortcite{wang2020cen}             & ResNet-101 & 39.59          & 19.32          & 29.46          \\
        TokenFusion  \shortcite{wang2022multimodal_token_fusion}  & MiT-B3 & 49.32          & 36.84          & 43.08          \\
        Reza et. al. \shortcite{reza2023robust}   & MiT-B4 & \textbf{52.82} & 36.72          & \underline{44.77}          \\
        MMANet \shortcite{wei2023mmanet}             & ResNet-50  & 44.93          & \textbf{42.75} & 43.84          \\
        HeMIS \shortcite{havaei2016hemis}              & ResNet-50  & 33.23          & 31.23          & 32.23          \\
        CMNeXt \shortcite{zhang2023cmnext}            & MiT-B4  & 51.19          & 5.26           & 28.23          \\
        RFNet  \shortcite{ding2021rfnet}             & ResNet-50 & 42.89          & 40.76          & 41.82          \\
        \textbf{MMP (Ours)} & MiT-B4  & \underline{52.04}    & \underline{41.08}    & \textbf{46.56} \\ \bottomrule
    \end{tabular}
    \caption{Performance (mIoU) comparison with existing methods on NYUDv2 dataset. RGB, Depth and Avg. columns report RGB only, Depth only and average performance respectively. Best and second-best results are shown as \textbf{bold} and \underline{underlined}, respectively.}
    \label{tab:comparison-nyu}
\end{table}

\begin{table}[t]
    \centering
    \setlength{\tabcolsep}{2pt}
    \begin{tabular}{cccccc}
    \toprule
    \multicolumn{2}{c}{Available Modality} & ViLT & \multicolumn{2}{c}{Missing Prompts \shortcite{lee2023map}} & MMP    \\
    Image               & Text               & \shortcite{kim2021vilt}    & Attention          & Input          & (Ours) \\
    \midrule
    100\%                        & 100\%                      & $\text{92.71}^\dag$ & 92.71          & 92.71          & \textbf{92.87} \\
    100\%                        & 30\%                       & 66.29               & 72.57          & \textbf{74.53} & 74.32    \\
    100\%                        & 0\%                        & $\text{23.70}^\dag$ & 67.70          & \textbf{68.10} & 66.06    \\
    65\%                         & 65\%                       & 69.25               & 78.09          & 79.08          & \textbf{80.28} \\
    30\%                         & 100\%                      & 76.66               & 86.05          & 86.18          & \textbf{87.71} \\
    0\%                          & 100\%                      & $\text{82.65}^\dag$ & 85.30          & 84.80          & \textbf{85.37}            \\ \bottomrule
    \end{tabular}
    \caption{Performance (accuracy) comparison for multimodal classification on UPMC Food-101 dataset. Available modality column shows the percentage of image and text modality available during inference. $\dag$ indicates that the results were generated using the available code from the authors.}
    \label{tab:comparison-food101}
\end{table}

\section{Experiments and Results}
\label{section:experiments}
In this section, we provide a comprehensive evaluation of our proposed method through detailed experiments on multimodal segmentation and classification tasks across five datasets. We compare our approach with established baseline methods that address missing modalities to assess its effectiveness and generalizability.

\subsection{Datasets}
\textbf{MCubeS dataset} \cite{Liang2022mcubes} contains 500 sets of images split into training, validation, and test sets containing 302, 96, and 102 sets of images respectively. It has 4 input modalities and per-pixel annotations for 20 material classes. \\
\textbf{NYUDv2 dataset} \cite{Silberman2012nyudv2} contains 1,449 aligned RGB-Depth image pairs, split into 795 for training and 654 for testing. Each images is 640 × 480 pixels and has annotation for 40 classes.\\
\textbf{FMB dataset} \cite{liu2023segmif} has 1,500 calibrated RGB-Infrared image pairs, with 1,220 for training and 280 for testing. It includes diverse scenes across various lighting and weather conditions, and offers per-pixel ground truth annotations for 14 classes.\\
\textbf{UPMC Food-101 dataset} \cite{wang2015food101} is a multimodal classification dataset containing image and text as input modalities. The dataset contains 90,704 image-text pairs divided into train, validation and test sets, and 101 food categories. \\
\textbf{CMU-MOSI dataset} \cite{zadeh2016mosi} is widely used for multimodal sentiment analysis and includes audio, visual, and text modalities. It contains a total of 2,199 samples, which are divided into training, validation, and test sets with 1,284, 229, and 686 samples respectively.\\
Complete details about each dataset are added in the supplementary materials of this paper.

\subsection{Implementation Details}
\label{sec:impl-details}
We use CMNeXt \cite{zhang2023cmnext} as the base model for multimodal segmentation, ViLT \cite{kim2021vilt} for multimodal classification, and multimodal transformer \cite{tsai2019mult} for multimodal sentiment analysis task. To evaluate missing modality performance, available modalities are provided while missing ones are set to zero for visual and audio data, and to empty strings for texts. For multimodal segmentation, we use a learning rate of $6 \times 10^{-5}$ with a polynomial scheduler, and apply OHEM cross-entropy loss for the FMB and MCubeS datasets, and cross-entropy loss for NYUDv2. We utilize AdamW \cite{loshchilov2017adamw}, with $\epsilon=10^{-8}$ and a weight decay of 0.01. For this task, we utilize the CMNeXt model pre-trained with modality dropout augmentation. We set the batch size to 4, and train the model for 500 epochs on MCubeS and NYUDv2 dataset. For the FMB dataset, we use a batch size of 2 and train for 120 epochs. For multimodal classification, we set the learning rate to $10^{-5}$, and utilize polynomial learning rate scheduler. We use cross-entropy loss, and AdamW optimizer with the same configuration as the multimodal segmentation task. Batch size is 16, and we train the model for 20 epochs. The remaining configurations of this task are the default settings from \cite{lee2023map}. For multimodal sentiment analysis we used the default settings from \cite{yu2021learning}. Further implementation details are added in the supplementary materials of this paper.

\subsection{Results on Multimodal Segmentation}

\begin{figure*}[ht]
    \centering
    \includegraphics[width=1.75\columnwidth]{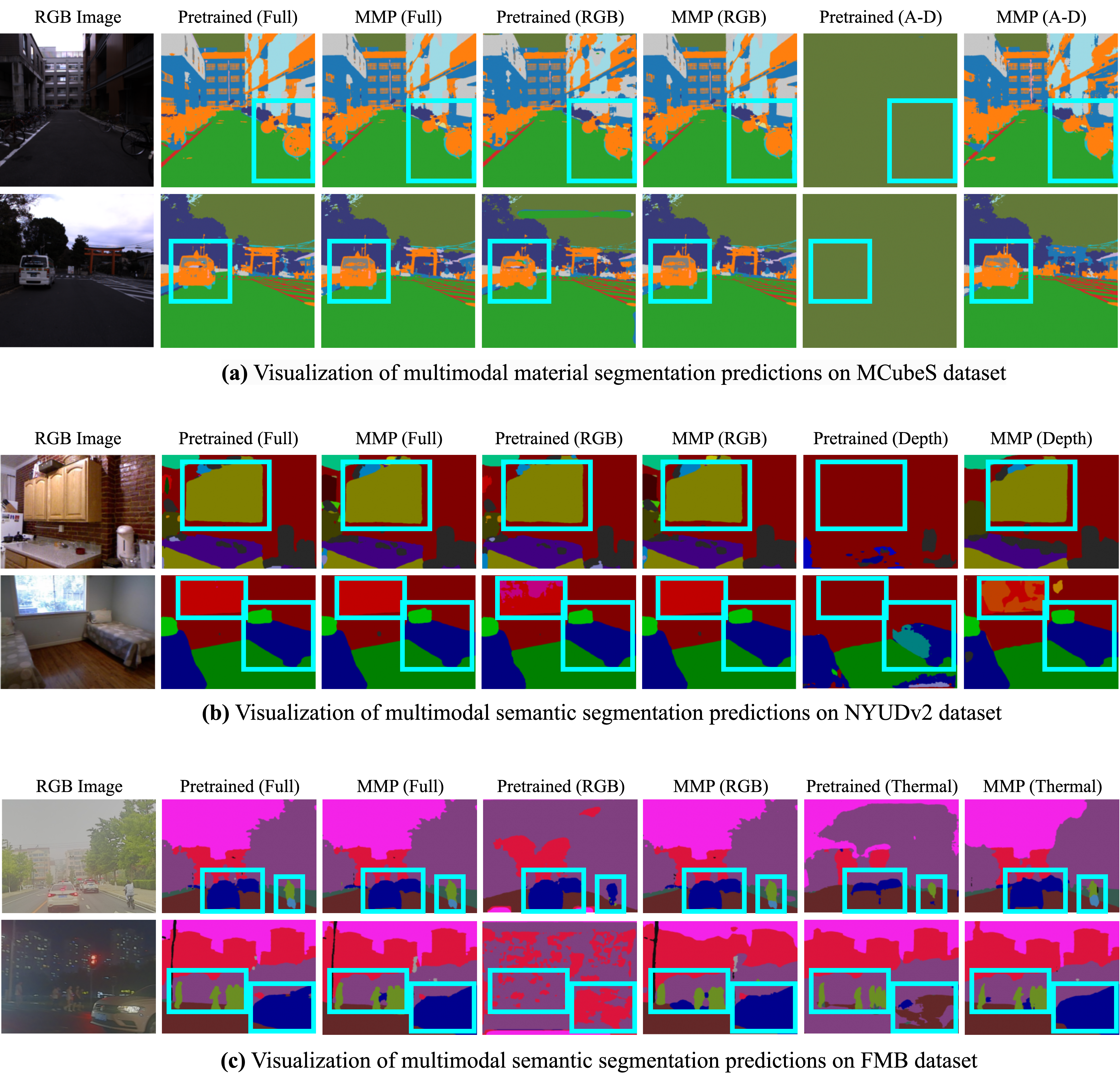} 
    \caption{Visualization of predicted segmentation maps for the Pretrained (CMNeXt) model and our MMP approach. Title above each image indicates the method name (available modalities). Blue boxes mark the areas where the differences are more prominent. A and D denote angle and degree of linear polarization, respectively.}
    \label{fig:visual-prediction}
\end{figure*}

\begin{table*}[t]
    \centering
    \begin{tabular}{@{}llcccccccc@{}}
    \toprule
    \multicolumn{1}{c}{Method} &
      \multicolumn{1}{c}{Backbone} &
      \multicolumn{2}{c}{Text-Visual-Audio} &
      \multicolumn{2}{c}{Visual-Audio} &
      \multicolumn{2}{c}{Audio} &
      \multicolumn{2}{c}{Average} \\
         &             & Acc   & F1    & Acc   & F1    & Acc   & F1    & Acc   & F1    \\ \midrule
    MulT \shortcite{tsai2019mult} & Transformer & \underline{79.57} & \underline{79.67} & 48.93 & 41.95 & 48.31 & 40.98 & 58.93 & 54.20 \\
    TFN \shortcite{zadeh2017tfn} & LSTM  & 73.90  & 73.40  & 42.23 & 25.07 & 42.23 & 25.07 & 52.78 & 41.18 \\
    LMF \shortcite{liu2018lmf}  & LSTM  & 76.40  & 75.70  & 43.29 & 27.61 & 42.23 & 25.07 & 53.97 & 42.79 \\
    Reza et. al. \shortcite{reza2023robust} &
      Transformer &
      \underline{79.57} &
      \underline{79.67} &
      \textbf{55.49} &
      \textbf{53.96} &
      \underline{50.00} &
      \underline{46.71} &
      \underline{61.68} &
      \underline{60.11} \\
    \textbf{MMP (Ours)} &
      Transformer &
      \textbf{80.03} &
      \textbf{80.04} &
      \underline{54.73} &
      \underline{52.24} &
      \textbf{55.03} &
      \textbf{53.98} &
      \textbf{63.26} &
      \textbf{62.08} \\ \bottomrule
    \end{tabular}
    \caption{Performance (binary accuracy and F1 score) 
    comparison with existing methods for multimodal sentiment analysis on CMU-MOSI dataset. Column names indicate available modalities. Best and second-best results are shown as \textbf{bold} and \underline{underlined}, respectively.}
    \label{tab:comparison-mosi}
\end{table*}

\begin{table}[t]
    \centering
    \setlength{\tabcolsep}{2.2pt}
    \begin{tabular}{@{}lcccc@{}}
    \toprule
    \multicolumn{1}{c}{Methods}        & RGB-Depth & RGB & Depth & Average        \\ \midrule
    Dropout                   & 51.12     & 48.80          & 29.79 & 43.23 \\
    Dropout + LP       & 51.31     & 51.08 & 35.48 & 45.95  \\
    Dropout + LP + Align & 52.84     & 50.73          & 40.60  & 48.05          \\
    Dropout + CA + Align               & \textbf{53.81}     & \textbf{52.04}          & \textbf{41.08} & \textbf{48.97} \\ \bottomrule
    \end{tabular}
    \caption{Ablation studies on NYUDv2 dataset. Modality dropout shows significant performance drop when RGB is missing. Performance increases as we apply linear projection (LP). Adding alignment loss (Align) improves performance further. Finally, replacing linear projection with cross attention (CA) shows overall best performance.}
    \label{tab:ablation-nyu}
\end{table}

We present a comparison of multimodal segmentation performance across the MCubeS, NYUDv2, and FMB datasets in Table~\ref{tab:comparison-mcubes}. All experiments were conducted using the CMNeXt \cite{zhang2023cmnext} model to ensure consistency and fairness. \textbf{Pretrained} column indicates the performance when we train the model without dropout augmentation and test the performance on different missing modality scenarios. \textbf{Modality dropout} column indicates the performance when we use dropout augmentation while training the model. \textbf{MMP} column indicates the performance when we utilize our modality projection approach to generate tokens for the missing modalities. 

Our findings indicate that the pretrained model experiences a notable performance drop when modalities get missing during test time. Although modality dropout improves performance in missing modality scenarios, it does not fully mitigate the performance drop. In contrast, our MMP approach utilizes the available modalities to generate tokens for missing ones, further enhancing robustness and performance across all datasets. Specifically, our MMP approach outperforms both the pretrained model and modality dropout in every missing modality scenario, demonstrating its effectiveness in compensating for missing modalities. 
The slightly reduced performance of the MMP approach when all modalities are available is due to the base model being pretrained with modality dropout, which has lower performance when all modalities are present.

We report experimental results for different baseline methods on NYUDv2 in Table~\ref{tab:comparison-nyu}. Compared to other methods, our MMP approach achieves superior average performance. When Depth is missing, MMP ranks second to Reza et. al. \shortcite{reza2023robust}, which learns adaptable layers for different modality combinations. The performance difference is minimal (-0.78\%), and MMP outperforms this method in both average (+1.79\%) and RGB-missing (+4.36\%) scenarios. This demonstrates that MMP has matched or exceeded the performance of this method while eliminating the computational overhead of adapting the model separately for each modality combination. When RGB is missing, MMP is the second-best to MMANet \shortcite{wei2023mmanet}. Notably, MMANet relies on a teacher model trained with all modalities, yet our MMP approach achieves better performance in other cases without the added complexity of training a separate teacher model.

\subsection{Visualization of Predictions}
To better demonstrate the impact of our approach, we visualize the predicted segmentation maps from the pretrained CMNeXt model and our MMP approach in Figure \ref{fig:visual-prediction}. For each dataset, we show the RGB image and predictions from the pretrained model and MMP approach with different available modalities (available modality names are shown in parentheses above each image). We highlight the impact of presence of the RGB modality due to its greater emphasis in the CMNeXt model's design.

In Figure \ref{fig:visual-prediction}a, we observe that the pretrained model struggles to detect the bikes and cars when the Angle of Linear Polarization (AoLP), Degree of Linear Polarization (DoLP), and Near-Infrared (NIR) modalities are missing. When RGB is missing along with NIR, and only AoLP and DoLP are available, the pretrained model fails to perform any segmentation. In contrast, our MMP approach successfully detects the bikes and cars even with missing modalities. On NYUDv2 dataset, as shown in Figure \ref{fig:visual-prediction}b, MMP demonstrates superior accuracy in detecting kitchen cabinets, counter-tops, windows, and beds compared to the pretrained model with missing modalities. In Figure \ref{fig:visual-prediction}c, the pretrained model fails to detect the car, bicyclist, and humans when modalities are missing. However, our MMP approach successfully detects all these objects in every scenario. In all examples, our MMP predictions are either better or comparable to the pretrained model with all modalities available. When modalities are missing, MMP consistently outperforms the pretrained model and closely matches the performance achieved with full input.

\subsection{Results on Multimodal Classification}
We further evaluate the effectiveness of our approach by performing a comparison against the missing-aware prompts method \cite{lee2023map} using the UPMC Food-101 \cite{wang2015food101} dataset. Experiments on multimodal classification task were conducted on ViLT \cite{kim2021vilt} as the base model. The results are detailed in Table~\ref{tab:comparison-food101}. Our MMP outperforms the prompting-based methods in most scenarios, achieving better overall results. Our performance shows a slight decrease in two cases, 0.21\% lower when 70\% of text is missing, and 2.04\% lower when no text is available. This is because prompting-based methods learn one set of prompts for each missing modality scenario and thus outperform MMP in certain modality combinations. Notably, our approach maintains strong performance across various missing modality scenarios without requiring training for every possible modality combination.

\subsection{Results on Multimodal Sentiment Analysis}
We evaluated our approach on the CMU-MOSI dataset \cite{zadeh2016mosi} for multimodal sentiment analysis, with results presented in Table~\ref{tab:comparison-mosi}. We utilized the multimodal transformer (MulT) \cite{tsai2019mult} as the base model for this task. Our findings indicate that when the text modality is present, missing audio, video, or both has little impact on performance, as noted in \citet{hazarika2022analyzing} and \citet{reza2023robust}. However, performance drops significantly without the text modality. Our MMP approach effectively compensates for this, providing a substantial improvement over the base MulT model. Specifically, we observe a 5.8\% improvement in accuracy and a 10.29\% increase in F1 score when text is missing. When only the audio modality is available, and both text and visual modalities are absent, MMP achieves larger improvements of 6.72\% in accuracy and 13\% in F1 score. MMP outperforms existing methods in both accuracy and F1 score across all scenarios, except when only text is missing, where it ranks second. The performance difference between MMP and the best-performing method (Reza et. al. \shortcite{reza2023robust}) in this case is minimal. However, MMP surpasses this method by a large margin in other scenarios, resulting in higher average accuracy and F1 scores overall.

\subsection{Ablation Studies}
To further investigate the contributions of various components in our proposed MMP approach, we conducted an ablation study on the NYUDv2 dataset. Table~\ref{tab:ablation-nyu} summarizes the results. We began with evaluating the performance of modality dropout, which serves as a baseline for comparison. Modality dropout shows significant performance drop when RGB is missing. Then we add a single linear layer as the projection function (LP) with modality dropout, to predict the tokens of the dropped modality using a linear combination of the available ones. This approach led to performance improvements across all scenarios. Next, we utilized an alignment loss (Align) objective with the linear projection and modality dropout. We observed further performance improvement in all cases, particularly when RGB was missing. Finally, we replaced the single linear layer with MMP's cross-attention based (CA) projection approach, combined with modality dropout and alignment loss. This final configuration achieved the highest performance among all the tested setups. Based on these experimental results, we argue that each component in our MMP module plays a critical role in enhancing the overall model performance in different missing modality scenarios.

\section{Conclusion} 
In this paper, we introduced Masked Modality Projection (MMP), a novel approach designed to enhance missing modality robustness of multimodal models. Our approach eliminates the need for training or adapting models for specific missing modality scenarios. We demonstrate that a single model can effectively handle any missing modality scenario and outperform current baselines. Thus it reduces both time and computational overhead. Experimental results across several baseline models and datasets validate that MMP significantly improves performance and robustness compared to existing baseline methods. Future work will focus on further refining MMP and exploring its applicability to other multimodal tasks and datasets. We believe that MMP offers an efficient and effective solution to the challenge of missing modalities. 

\medskip 
\noindent \textbf{Acknowledgment:}
This work is supported in part by AFOSR award FA9550-21-1-0330 and an Amazon Gift award. This work used Indiana Jetstream2 through allocation CIS220128 from the ACCESS program supported by NSF grants 2138259, 2138286, 2138307, 2137603, and 2138296.

\newpage 
\bibliography{main}

\include{supp}

\end{document}

%% file: supp.tex
\setcounter{page}{1}

\newcommand{\beginsupplement}{%
        \setcounter{table}{0}
        \renewcommand{\thetable}{S\arabic{table}}%
        \setcounter{figure}{0}
        \renewcommand{\thefigure}{S\arabic{figure}}%
        \setcounter{section}{0}
        \renewcommand{\thesection}{S\arabic{section}}%
     }

\beginsupplement

\twocolumn[
\begin{center}
    {\LARGE  \bf Supplementary Material \\ \medskip 
    MMP: Towards Robust Multi-Modal Learning with Masked Modality Projection}
\end{center}
]

\begin{figure*}[ht]
    \centering
    \includegraphics[width=2\columnwidth]{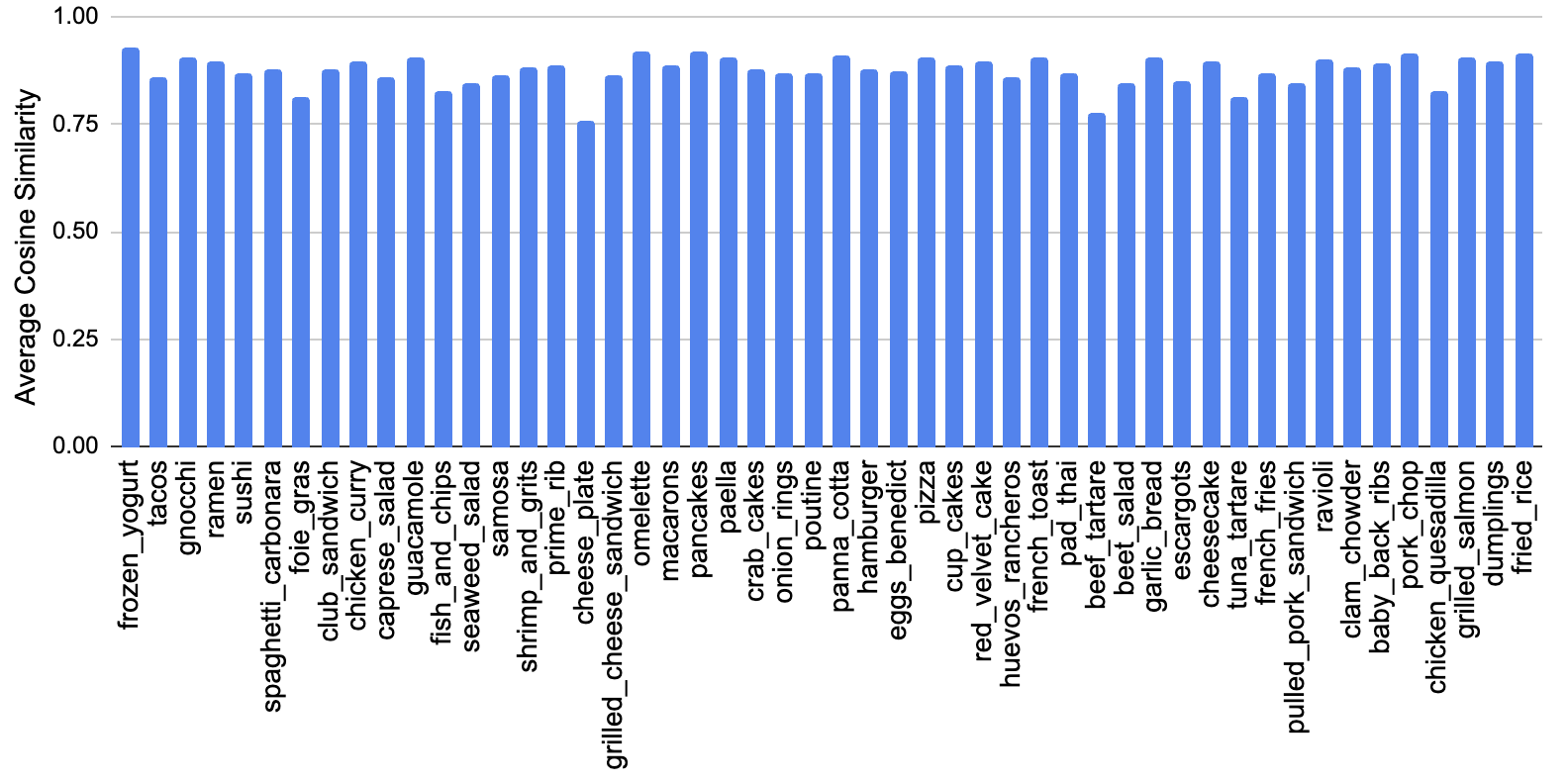} 
    \caption{Average cosine similarity between model predictions with real and projected tokens on UPMC Food-101 dataset when image is missing. We substitute the missing modality tokens with the projected tokens.}
    \label{fig:cosine-food101-image-missing}
\end{figure*}

\begin{figure*}[ht]
    \centering
    \includegraphics[width=2\columnwidth]{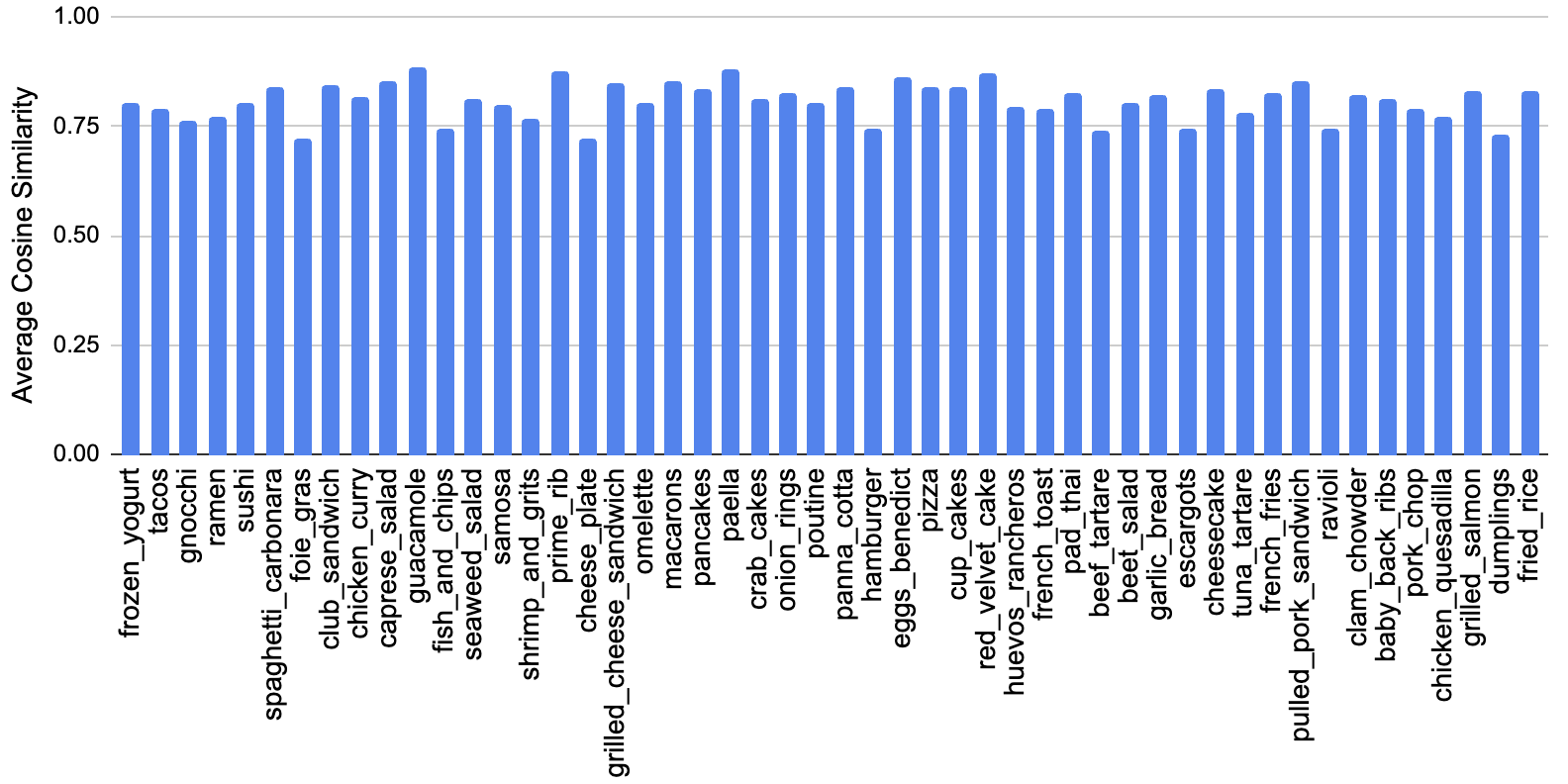} 
    \caption{Average cosine similarity between model predictions with real and projected tokens on UPMC Food-101 dataset when text is missing. We substitute the missing modality tokens with the projected tokens.}
    \label{fig:cosine-food101-text-missing}
\end{figure*}
\setcounter{section}{0}
\section{Datasets}
\label{supp-datasets}

In this section, we provide an overview of the datasets used in our experiments. The motivation for selecting these datasets lies in their popularity and widespread usage in the field. These datasets cover a broad range of tasks, including multimodal segmentation, classification, and sentiment analysis. We utilize these datasets to include both homogeneous and heterogeneous data types, aiming to conduct a comprehensive evaluation of the effectiveness of our approach across different modality types.

\textbf{MCubeS dataset} \cite{Liang2022mcubes} is a multi-modal dataset featuring four distinct input modalities: RGB, Angle of Linear Polarization (AoLP), Degree of Linear Polarization (DoLP), and Near-Infrared (NIR). The dataset is organized into three subsets: 302 sets of images for training, 96 sets of images for validation, and 102 sets of images for testing. The dataset is annotated with per-pixel labels across 20 different material classes. This detailed annotation allows for comprehensive analysis and modeling of material properties under diverse imaging conditions.

\textbf{NYUDv2 dataset} \cite{Silberman2012nyudv2} comprises 1,449 aligned RGB-Depth image pairs, which are divided into 795 pairs for training and 654 pairs for testing. Each image pair has a resolution of 640 × 480 pixels and includes detailed annotations for 40 distinct classes. For our experiments, we utilize HHA-encoded images instead of raw depth maps following recent studies \cite{zhang2023cmnext, reza2023robust}.

\textbf{FMB dataset} \cite{liu2023segmif} is a comprehensive dataset consisting of 1,500 pairs of calibrated RGB-Infrared images. The dataset is divided into 1,220 image pairs for training and 280 image pairs for testing. It covers a broad spectrum of scenes, including various lighting and weather conditions such as the Tyndall effect, rain, fog, and intense lighting. It has per-pixel ground truth annotation for 14 classes.

\textbf{UPMC Food-101 dataset} \cite{wang2015food101} is widely used for multimodal classification. It contains 90,704 image-text pairs, which are split into training, validation, and test sets. The dataset features annotations for 101 distinct classes, consistent with those in the ETHZ Food-101 dataset \cite{bossard2014food}. 

\textbf{CMU-MOSI dataset} \cite{zadeh2016mosi} is a popular dataset for multimodal sentiment analysis. It consists of 2,199 samples, each including audio, visual, and text modalities. The dataset is split into training, validation, and test sets with 1,284, 229, and 686 samples respectively, and includes sentiment annotations for each sample.

\section{Implementation Details}
\label{sec:impl-details-supplementary}
We use Python\footnote{https://www.python.org/} 3.8.18 and PyTorch\footnote{https://pytorch.org/} 2.1.1 for multimodal segmentation and classification, and Python 3.10.14 and Pytorch 2.4.0+cu121 for sentiment analysis task. Experiments are conducted using two NVIDIA RTX 2080 Ti GPUs with 16G memory. The code is configured with fixed seeds to ensure the results can be replicated.

To assess missing modality performance, we provide the available modalities to the model while setting the missing modalities to zero for visual and audio data, and to empty strings for texts. 

\subsection{Multimodal segmentation}
We use CMNeXt \cite{zhang2023cmnext} as the base model for multimodal segmentation. We use their publicly available code\footnote{https://github.com/jamycheung/DELIVER/} to train the base model with dropout augmentation. Then we use the resulting weights to initialize our models. The learning rate is set to $6 \times 10^{-5}$ and polynomial learning rate scheduler is applied
with power = 0.9. The first 10 epochs are warm-up epochs and the learning rate is set to 0.1 times the original rate. We use OHEM cross-entropy loss for FMB and MCubeS datasets, and cross-entropy loss for NYUDv2 dataset. For the optimizer, we utilize AdamW \cite{loshchilov2017adamw} with $\epsilon=10^{-8}$ and weight decay = 0.01. We train the model with a batch size of 4 for 500 epochs on the MCubeS and NYUDv2 datasets. For the FMB dataset, we use a batch size of 2 and train for 120 epochs.

\begin{figure}[ht]
    \centering
    \includegraphics[width=1\columnwidth]{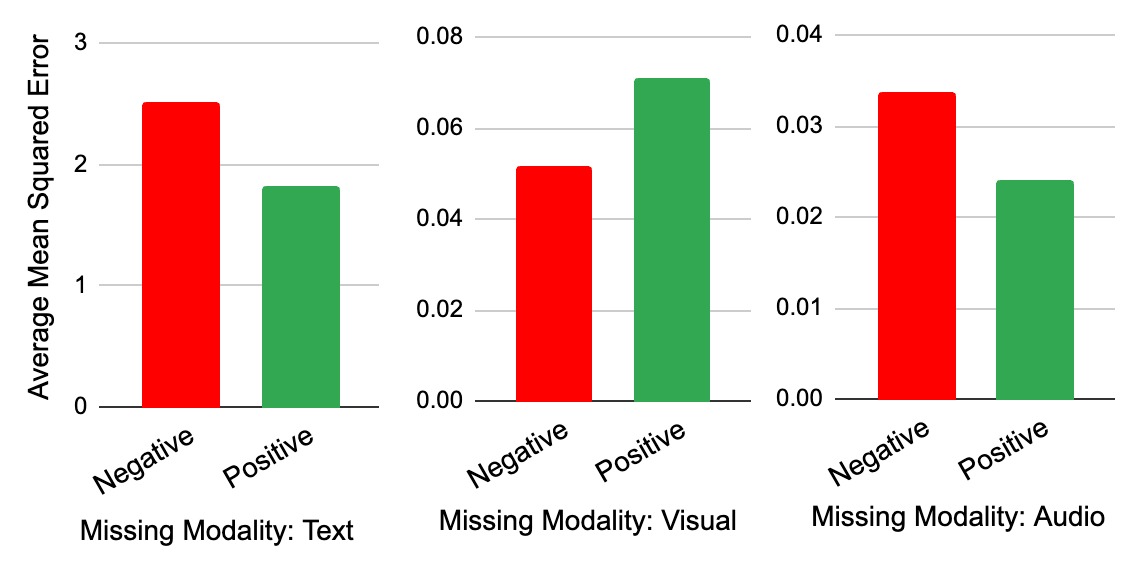} 
    \caption{Average Mean Squared Error (MSE) between model predictions with real and projected tokens on CMU-MOSI dataset.}
    \label{fig:mse-mosi}
\end{figure}

\textbf{MCubeS dataset:} We adopt the data pre-processing and augmentation techniques outlined in \citet{zhang2023cmnext}. The MiT-B2 backbone from \citet{xie2021segformer} is utilized for this dataset. For training, the input images are resized to $512 \times 512$, while during testing, they are set to $1024 \times 1024$. The results are reported based on single-scale performance using predicted segmentation maps at $1024 \times 1024$. %

\textbf{NYUDv2 dataset:} We utilize HHA-encoded images rather than raw depth maps following SA-Gate \cite{chen2020bi} and CMNeXt \cite{zhang2023cmnext}. The preprocessed dataset can be downloaded from the SA-Gate repository\footnote{https://github.com/charlesCXK/ \\ RGBD\_Semantic\_Segmentation\_PyTorch/}. Both RGB and HHA images are resized to $640 \times 480$ pixels for both training and testing. Following the recommendation in the CMNeXt paper, we employ MiT-B4 backbone from \citet{xie2021segformer}. %

\textbf{FMB dataset:} We follow the data pre-processing and augmentations used by \citet{liu2023segmif}, and MiT-B3 from \citet{xie2021segformer} is used as the backbone. We set the input image resolution to $800 \times 600$ during both training and testing. %

\subsection{Multimodal classification}
\textbf{UPMC Food-101 dataset:} For multimodal classification task, we use ViLT \cite{kim2021vilt} as our base model. We set the learning rate to $10^{-5}$, and polynomial learning rate scheduler is applied with power = 0.9, ratio = 0.1, and the first 2500 steps as the warm-up. We use cross-entropy loss, and AdamW optimizer with a weight decay of 0.02. Batch size is 16, and we train the model for 20 epochs. The remaining configurations for this task are the same as \citet{lee2023map}. 

\subsection{Multimodal sentiment analysis}
\textbf{CMU-MOSI dataset:} For multimodal sentiment analysis, we use multimodal transformer \cite{tsai2019mult} from this publicly available code\footnote{https://github.com/thuiar/MMSA/}, and use the default configurations from \citet{yu2021learning}.

\subsection{Reproducibility Statement}
We made our source code and pretrained models available at this anonymous link\footnote{https://drive.google.com/drive/folders/155IsgLD88-Dt6Q9DJCKCENuDF7vjI9jI} to ensure the reproducibility of our results, and to facilitate the community in building upon our work. A comprehensive README.md file is added containing detailed instructions to set up environment, execute code, and reproduce the results. The main experimental results presented in this paper can be reproduced using the provided scripts, pretrained model weights, and instructions. 

\section{Projected and Real Tokens Alignment}
In this section, we examine the alignment between the real tokens of a modality and the projected tokens generated from the available modalities using our MMP approach. The alignment of these tokens is crucial. This ensures that the model can effectively predict and substitute the missing modality tokens from the available modalities. We employed a smooth L1 loss to align the projected tokens with the real tokens during training as described in Section \ref{sec:method-alignment}. This alignment loss minimizes the discrepancy between the projected tokens and the original tokens. To evaluate the effectiveness of this alignment, we employ cosine similarity for multimodal classification on the UPMC Food-101 dataset and Mean Squared Error (MSE) for multimodal sentiment analysis on the CMU-MOSI dataset. These metrics help us visualize the degree to which the model's predictions align when using real tokens versus projected tokens during testing.

\subsection{UPMC Food-101 dataset}
The first analysis is conducted on the UPMC Food-101 dataset, as illustrated in Figures \ref{fig:cosine-food101-image-missing} and \ref{fig:cosine-food101-text-missing}. These figures show the cosine similarity between the model's predictions using real tokens and projected tokens during testing. We show the cosine similarity of the first 50 out of 101 classes. Figure \ref{fig:cosine-food101-image-missing} visualizes the scenario where the image modality is missing, and its tokens are replaced with projected tokens generated from the available text modality. The results show high cosine similarity, with all similarity scores exceeding 0.75 and most classes achieving scores above 0.8.

Similarly, Figure \ref{fig:cosine-food101-text-missing} shows the cosine similarity when the text modality is missing. We see similar pattern here, with scores consistently above 0.7 and mostly above 0.8. This indicates that the MMP approach effectively projects the available modality tokens to estimate the missing ones, aligning these tokens in a manner that ensures stability in the model's predictions.

\subsection{CMU-MOSI dataset}
We extend this analysis to the CMU-MOSI dataset. Figure \ref{fig:mse-mosi} illustrates the average Mean Squared Error (MSE) for each modality. The results show that the audio modality exhibits the lowest MSE, reflecting the highest alignment between real and projected tokens. This is followed by the visual modality, which also demonstrates low MSE. The text modality shows a higher MSE, though it remains below 3. This observation is consistent with related work, which has shown that the text modality often has the most significant impact when missing, making it more challenging to project accurately.

\section{Evaluation Metrics}
\label{sec:evaluation_metrics}

In this section, we describe the evaluation metrics used to assess the performance of our MMP approach across three different tasks. We utilize mean Intersection over Union (mIoU) for segmentation, accuracy for classification, and binary accuracy and F1-score for sentiment analysis. These metrics are standard in their respective fields, and it is common practice to use them for benchmarking against related work.

\subsection{Multimodal Segmentation}
For the multimodal segmentation task, we use the mean Intersection over Union (mIoU) as the evaluation metric. mIoU is a standard metric for evaluating semantic segmentation performance. It calculates the average overlap between the predicted segmentation and the ground truth across all classes. This provides a comprehensive evaluation by considering both false positives and false negatives, and capturing the spatial overlap between the predicted and true segmentations.

\subsection{Multimodal Classification}
For the multimodal classification task, we employ accuracy as our evaluation metric. Accuracy measures the proportion of correct predictions made by the model out of all predictions. This is a straightforward and intuitive metric that provides an overall performance measure of the classification model. 

\subsection{Multimodal Sentiment Analysis}
For the multimodal sentiment analysis task, we utilize both accuracy and F1-score to evaluate the model's performance. F1-score is the mean of precision and recall, providing a balance between the two. We use the binary metrics in this task, which only consider the Negative and Positive classes, following the common approach in related work.